\documentclass[conference]{IEEEtran}

\usepackage[ngerman,english]{babel}
\usepackage{color}
\usepackage[T1]{fontenc}
\usepackage{graphicx}
\usepackage[utf8]{inputenc}
\usepackage{listings}
\usepackage{xspace}
\usepackage{amsmath,amssymb}
\usepackage{tabularx,multirow}
\usepackage{footmisc}
\usepackage{etoolbox}
\usepackage{longtable}
\usepackage[listofformat=parens, justification=centering, subrefformat=subparens]{subfig}  
\usepackage[table]{xcolor}
\usepackage[export]{adjustbox}
\usepackage{import}
\usepackage{subfiles}
\usepackage{comment}
\usepackage{longtable}
\usepackage{listings}
\usepackage{nicematrix}

\usepackage[breaklinks=true,colorlinks,bookmarks=true,hypertexnames=false]{hyperref}  

\usepackage{todonotes}

\newcommand{\secref}[1]{Section~\ref{#1}\xspace}
\newcommand{\figref}[1]{Figure~\ref{#1}\xspace}

\newcommand{\tabref}[1]{Table~\ref{#1}\xspace}

\newcommand{\ie}{\textit{i.\,e.}\xspace}

\newcommand\Caption[3][]{%
  \caption[#2]{%
    \ifstrempty{#1}{}{\label{#1}}%
    \textsc{#2}. \it #3%
  }%
}

\newcommand\T{\ensuremath{^\mathrm{T}}}

\newcolumntype{C}{X<{\centering}}

\title{Biased Binary Attribute Classifiers Ignore the Majority Classes}

\iffalse
\author{Anonymous Swiss Conference on Data Science submission}

\else
\author{\IEEEauthorblockN{%
  Xinyi Zhang,\IEEEauthorrefmark{1}
  Johanna Sophie Bieri,\IEEEauthorrefmark{2}
  Manuel G\"unther\IEEEauthorrefmark{3}
}
\IEEEauthorblockA{%
  Department of Informatics,
  University of Zurich\\
  Andreasstrasse 15,
  8050 Zurich, Switzerland \\
  Email: \IEEEauthorrefmark{1}xinyi.zhang@uzh.ch,
  \IEEEauthorrefmark{2}johanna.bieri@uzh.ch,
  \IEEEauthorrefmark{3}manuel.guenther@uzh.ch
}
}
\fi

\usepackage{fancyhdr}
\begin{document}
\thispagestyle{empty}
\maketitle

{
  \chead{\footnotesize This is a pre-print of the original paper accepted for publication at the \href{https://sds2024.ch/}{Swiss Conference on Data Science (SDS)} 2024.}
  \lhead{}
  \thispagestyle{fancy}
}

\def\citep#1{\cite{#1}}

\begin{abstract}

To visualize the regions of interest that classifiers base their decisions on, different Class Activation Mapping (CAM) methods have been developed.
However, all of these techniques target categorical classifiers only, though most real-world tasks are binary classification.
In this paper, we extend gradient-based CAM techniques to work with binary classifiers and visualize the active regions for binary facial attribute classifiers.
When training an unbalanced binary classifier on an imbalanced dataset, it is well-known that the majority class, \ie the class with many training samples, is mostly predicted much better than minority class with few training instances.
In our experiments on the CelebA dataset, we verify these results, when training an unbalanced classifier to extract 40 facial attributes simultaneously.
One would expect that the biased classifier has learned to extract features mainly for the majority classes and that the proportional energy of the activations mainly reside in certain specific regions of the image where the attribute is located.
However, we find very little regular activation for samples of majority classes, while the active regions for minority classes seem mostly reasonable and overlap with our expectations.
These results suggest that biased classifiers mainly rely on bias activation for majority classes.
When training a balanced classifier on the imbalanced data by employing attribute-specific class weights, majority and minority classes are classified similarly well and show expected activations for almost all attributes.
\end{abstract}


\section{Introduction}

Binary classification tasks are prevalent in many applications.
Unfortunately, many binary classification datasets are highly imbalanced, i.e., one of the two classes appears much more often than the other.
When training a classifier on such a biased dataset, it has been shown that the classifier mainly learns the majority class\footnote{We make use of the terms \emph{majority} and \emph{minority} class to refer to the classes with large and small amounts of training samples, respectively.} and predicts poorly on the minority class \citep{rudd2016moon}.
Our experiments validate this behavior.

Since the classifier sees many more samples of the majority class during training, one would expect that it learns the features required to classify this class very well.
To assess whether such an assumption actually holds, we make use of techniques for interpretability.
Particularly, visualization techniques such as the family of Class Activation Mapping (CAM) methods \citep{zhou2016cam} have been used to analyze the input regions of images that contribute most to the classification.
Many of these techniques make use of the network gradients \citep{selvaraju2017gradcam,chattopadhay2018grad-cam++,draelos2020hirescam} to improve the predictions.
Yet, these CAM techniques are designed for categorical classifiers, \ie, where more than two classes are predicted, and most of them focus on activations resulting from SoftMax.
For most binary classifiers, however, only one output is available that presents the prediction for the positive class.
Therefore, most categorical classifiers can only highlight the activation of the \emph{positive} class, whereas for binary classifiers it is more important to highlight the \emph{predicted} class.
To achieve this, a small modification is applied to gradient-based CAM techniques, which we will present in this paper.

Since the CelebA dataset \citep{liu2015deep} contains facial attributes with different severity of imbalance, this dataset provides a perfect testbed for our experiments.
Using our new technique, we visualize facial attributes extracted by the state-of-the-art Alignment-Free Facial Attribute Classifier (AFFACT) \citep{guenther2017affact}.
This classifier is trained on the raw CelebA dataset, \ie, without taking its bias into account.
For highly imbalanced attributes, one would expect that the classifier learns to extract the most important features from the majority classes, while minority classes contribute only little to the learned features.
Surprisingly, our experiments show the exact opposite: The classification of the majority class is based on the corners of the images or the bias neuron of the final layer.
Even worse, since AFFACT learns to predict all attributes simultaneously, this kind of behavior is even propagated to the mostly balanced attributes.

One way of fighting against the bias in a classifier is by artificially balancing the training data.
This can be done in various ways, here we select one promising approach that can handle multiple outputs simultaneously.
Particularly, by combining the training method of AFFACT and the debiasing technique of the Mixed Objective Optimization Network (MOON) from \cite{rudd2016moon}, we show that we can train a classifier that is better suited for classifying minority classes, and that this classifier learns to base its predictions on the relevant parts of the images, for both the majority and minority classes.

\section{Related Work}

Many classification tasks throughout the field of research are binary, \ie, the model needs to discern between two classes.
Examples are identifying man and women \citep{lin2016human}, spam email \citep{mansoor2021comprehensive}, malware \citep{rudd2017survey}, or skin cancer \citep{esteva2017dermatologist}.
While some of these tasks are balanced, \ie, both positive and negative classes appear similarly often, many of them are highly imbalanced such that one class appears much more often than the other \citep{kumar2022review}.

\begin{figure*} [t]
    \centering
    \includegraphics[width=.9\textwidth]{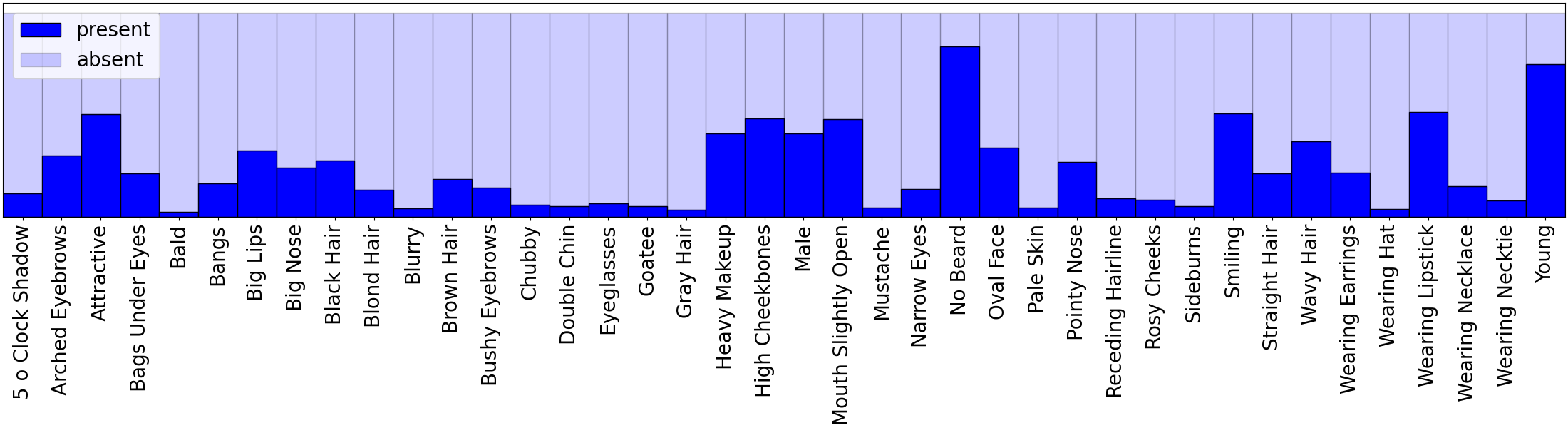}
    \Caption[fig:attributes]{Distribution of Attributes}{This figure shows the distribution of the binary facial attributes throughout the CelebA dataset, indicating its large imbalance for some attributes (replicated from \cite{rudd2016moon}).}
\end{figure*}

\subsection{Facial Attribute Classification}
One specific task that includes both balanced and unbalanced binary classification tasks is facial attribute prediction.
Particularly, the CelebA dataset \citep{liu2015deep} contains 40 binary facial attributes, some of which are balanced (such as \texttt{Attractive}) and some are highly imbalanced (such as \texttt{Chubby}), as can be seen in \figref{fig:attributes}.
Generally, there exist two approaches for facial attribute classification, single-label learning methods which make predictions for each attribute separately, and multi-label learning methods that predict facial attributes concurrently \citep{mao2020deep}.
While early work \citep{kumar2009attribute,liu2015deep,rozsa2016attributes,zhong2016leveraging} relied on single-attribute classifiers, it was realized that combined approaches \citep{rudd2016moon,hand2017attributes,guenther2017affact,zhuang2018multilabel,rozsa2019accuracy} can leverage from correlations between attributes.
While many approaches to jointly classify facial attributes simply ignore the imbalanced nature of some attributes \citep{hand2017attributes,zhuang2018multilabel,guenther2017affact}, several approaches have been made to provide more balanced and less biased attribute classifiers \citep{rudd2016moon,huang2016learning,kalayeh2017improving}.

\subsection{Class Activation Mapping}
To be able to shed some light into the interpretability of machine learning models, there exist several techniques of visualizing the importance of local regions in the input for the final classification \citep{linardatos2020explainable}.
One particular class of methods is based on Class Activation Mapping (CAM) \citep{zhou2016cam}, for which several extensions have been proposed \citep{selvaraju2017gradcam,chattopadhay2018grad-cam++,draelos2020hirescam,wang2020score-cam,jiang2021layercam}.
Many of these methods rely on network gradients and are generally designed to visualize categorical classifiers.
Compared with traditional Grad-CAM \cite{selvaraju2017gradcam}, Grad-CAM++ \citep{chattopadhay2018grad-cam++} produces a visual explanation for the class label under consideration by using a weighted mixture of the positive partial derivatives of the last convolutional layer feature maps with respect to a certain class score as weights.
Because of its gradient averaging step, Grad-CAM occasionally experiences problems with inaccurate positions.
To solve this problem, HiResCAM \citep{draelos2020hirescam} multiplies the activations with the gradients, which can provably guarantee faithfulness for certain models.
Element-wise Grad-CAM \citep{gildenblat2021pytorchcam} is another variant of the Grad-CAM, which multiplies the activations element-wise with the gradients first and then applies a ReLU operation before summing.

In preliminary work \cite{wu2023consistency}, we have performed first approaches of visualizing facial attribute classification.
However, we only visualized a single balanced attribute extracted with a balanced attribute classifier \citep{rudd2016moon} and a single non gradient-based visualization technique \citep{wang2020score-cam}.
This work is based on \cite{bieri2023bachelor}.

\section{Approach}

The aim of this paper is to highlight the properties of binary classifiers when trained on imbalanced and balanced datasets.
First, we adapt gradient-based CAM techniques to work with binary classifiers.
To compare unbalanced and balanced models, we train a balanced model on the imbalanced CelebA dataset.
Finally, we evaluate our attribute classifiers by defining regions of the image for each attribute where we would expect the classifier to extract information from, and use these regions to evaluate the interpretability of our classifiers.

\subsection{Visualizing Binary Classifiers}

Binary classifiers often use a single output neuron for predicting the presence of the positive class.
Usually, a logit:
\begin{equation}
  \label{eq:logit}
  z=\vec w\T \vec \varphi +b
\end{equation}
is computed for a given deep feature $\vec \varphi$, a learnable fully-connected weight vector $\vec w$ and a bias neuron $b$.
Afterward, the logit is transformed to a probability using the logistic activation function:
\begin{equation}
  \label{eq:probability}
  y = 1/(1+e^{-z})\,.
\end{equation}
During inference, the probability $y$ is thresholded at 0.5 to obtain a yes/no answer.
Instead, one could also threshold the logit $z$ at 0 to achieve the same result.

Class Activation Mapping (CAM) techniques \citep{zhou2016cam} only work with categorical classifiers, where the contribution for a certain class $c$ shall be predicted.
This map estimates the activation at the given spatial location $(i,j)$ by computing a weighted average over the feature map $f_k(i,j)$:
\begin{equation}
  \label{eq:cam}
  A^c(i,j) = \sum_{k} \alpha^c_kf_k(i,j)
\end{equation}
where $\alpha_k^c$ is the weight for channel $k$ when classifying class $c$.
This activation map $A$ is then rescaled to the input image dimensions, commonly using bilinear interpolation.
Typically, gradient-based CAM techniques compute these weights by back-propagating the output $y^c$ to the feature map and computing some aggregate of these across locations \citep{selvaraju2017gradcam}:
\begin{equation}
  \label{eq:gradcam}
  \alpha^c_k = \sum_{(i,j)} \frac{\partial y^c}{\partial f_k(i,j)}\,.
\end{equation}
The interpretation of this gradient is: In which direction would the feature map need to change in order to increase the probability that class $c$ is predicted?

We apply a similar interpretation for computing the weights for binary classifiers.
Here, we take one assumption that includes all binary classifiers trained with logistic activation, but also embraces  other loss functions such as the one proposed by \cite{rudd2016moon}: We threshold the logit score $z$ at 0.
Thus, when the classifier predicts the negative class, stronger negative logit values will increase the prediction of the negative class.
Hence, for a binary classifier, we can compute the weight $\alpha_k$ (note that we have only one output here) as:
\begin{equation}
  \label{eq:gradcam-binary}
  \alpha_k = \sum_{(i,j)} \frac{\partial |z|}{\partial f_k(i,j)} = \sum_{(i,j)} \begin{cases} \frac{\partial z}{\partial f_k(i,j)} & \text{if } z>0 \\[1ex] \frac{\partial (-z)}{\partial f_k(i,j)} & \text{else} \end{cases}
\end{equation}
This technique can be applied to several gradient-based CAM methods.
In our experiments, we make use of the visualization techniques implemented by \cite{gildenblat2021pytorchcam}, which allows us to specify the loss function $|z|$ according to \eqref{eq:gradcam-binary} for various CAM techniques.
We also utilize their default functionality to overlay activations on images.
Finally, we show the average activation over several images with the same predictions.

Obviously, there exists the possibility to train binary classifiers with two output neurons and softmax, in which case standard grad-CAM techniques can be applied.
However, these techniques fail to visualize effects in pre-trained binary classifiers that only have a single output neuron.

\subsection{Balancing Facial Attribute Classifiers}
\label{sec:balancing}

For training our balanced facial attribute classifier, several options are possible \cite{kumar2022review}.
Here, we selected to combine the two approaches that we proposed in \cite{rudd2016moon} and \cite{guenther2017affact} to arrive at the balanced AFFACT-b network, other balancing techniques shall be investigated in future work.
To be comparable to the original unbalanced AFFACT network from \cite{guenther2017affact}, which we term AFFACT-u, we use the exact same network, parameters and training schedules.
Particularly, we employ a pre-trained ResNet-50 network on ImageNet, which we extend with an additional logit layer to predict 40 facial attributes simultaneously.
We apply the same data augmentation as in \cite{guenther2017affact}.

For balancing the distributions of attributes, we use the Euclidean loss function \citep{rudd2016moon} averaged over $N$ training samples:
\begin{equation}
  \label{eq:moon-loss}
  \mathcal J_w = \frac1N\sum_{n=1}^N\sum\limits_{m=1}^M w_m(t_{nm}) \cdot (z_{nm} - t_{nm})^2
\end{equation}
where $m$ represents one of the $M=40$ different attributes, $t_{nm} \in \{+1,-1\}$ the target label of that attribute, and $z_{nm}$ the prediction of attribute $m$ for sample $n$.
For each attribute, we count the probability $p_m$ that a training sample comes from the positive class, and we compute the weight $w_m(t_{nm})$ for the two cases in order to balance the distributions of both classes:
\begin{equation}\begin{aligned}
  \label{eq:moon-weights}\hspace*{-1em}
  p_m &= \frac1N \sum_n\delta_{1,t_{nm}} \\
  w_m(+1) &= \begin{cases} 1 & \text{if } p_m > 0.5\\ \frac{1-p_m}{p_m} & \text{else}\end{cases} \\
  w_m(-1) &= \begin{cases} \frac{p_m}{1-p_m} & \text{if } p_m > 0.5\\ 1 & \text{else}\end{cases}
\end{aligned}\end{equation}
where $\delta$ is the Kronecker delta.
When assuming a balanced target distribution of classes per attribute, \eqref{eq:moon-weights} results in the exact same weights as we proposed in \cite{rudd2016moon}.
While previously, we used the weight as a probability to sample whether the loss is applied \cite{rudd2016moon}, here we directly apply the weight as a multiplicative factor in \eqref{eq:moon-loss}.

\subsection{Selecting Frontal Test Images}

Since the main focus of our work is the visualization of facial attributes, we only select frontal faces so that a simple aggregation of samples is possible without non-frontal faces disturbing our average CAM results.
Also, since most images in CelebA are frontal, we want to exclude random effects arising from non-frontal images that might not have a good representation in the trained models.

For this purpose, we used a simple heuristic on the hand-labeled facial landmarks of the CelebA dataset.
Particularly, we computed distance of the nose landmark from the line connecting the center of the mouth corners with the center of the eyes, relative to the distance between eyes center and mouth center.
Only faces that had a relative distance smaller than 0.1 were considered as frontal, any other face was excluded from our evaluation.
Using this filtering, we obtain 10'458 frontal faces out of the 19'962 CelebA test samples.

\subsection{Evaluation Metrics}

Since we evaluate balanced and unbalanced binary attribute classifiers, we need to adapt our evaluation technique accordingly.
Therefore, we compute False Negative Rates (FNR) and False Positive Rates (FPR) separately per attribute to show which of the two classes is predicted well.
For the unbalanced network, we expect the majority class to be classified well, while the minority class likely has higher error rates.

For evaluating the CAM visualizations, we rely on proportional energy \citep{wang2020score-cam}, which counts how much of the visualization energy is inside a certain binary mask $B$ that we define separately for each attribute.
Please refer to the supplemental material\footnote{Supplemental material as well as code for our evaluation can be found in our source code package \url{http://github.com/AIML-IfI/attribute-cam}.} and \cite{bieri2023bachelor} for more details on how we selected these masks.
The proportional energy is defined as:
\begin{equation}
  \label{eq:prop-energy}
  E = \frac{\sum\limits_{(i,j)} B(i,j)\cdot A^*(i,j)}{\sum\limits_{(i,j)} A^*(i,j)}
\end{equation}
where $A^*$ represents the activations $A$ from \eqref{eq:cam} scaled to input resolution.
We set $E=0$ when there is no activation $A^*(i,j)$ at any location and, therewith, the denominator vanishes.

\section{Experiments}

\subsection{Classification Errors}

\newcommand\head[2]{\multicolumn{2}{c}{\tiny\bf{#1} -- #2\,\%}}
\newcommand\icb[1]{\includegraphics[width=.08\textwidth]{images/balanced/grad-cam/pr=-1/#1} & \includegraphics[width=.08\textwidth]{images/balanced/grad-cam/pr=1/#1}}
\newcommand\icu[1]{\includegraphics[width=.08\textwidth]{images/unbalanced/grad-cam/pr=-1/#1} & \includegraphics[width=.08\textwidth]{images/unbalanced/grad-cam/pr=1/#1}}
\newcommand\headelementwise[2]{\multicolumn{2}{c}{\tiny\bf{#1} -- #2\,\%}}
\newcommand\icbelementwise[1]{\includegraphics[width=.08\textwidth]{images/balanced/grad-cam-elementwise/test-pr=-1/#1} & \includegraphics[width=.08\textwidth]{images/balanced/grad-cam-elementwise/test-pr=1/#1}}
\newcommand\icuelementwise[1]{\includegraphics[width=.08\textwidth]{images/unbalanced/grad-cam-elementwise/test-pr=-1/#1} & \includegraphics[width=.08\textwidth]{images/unbalanced/grad-cam-elementwise/test-pr=1/#1}}
\newcommand\headxcam[2]{\multicolumn{2}{c}{\tiny\bf{#1} -- #2\,\%}}
\newcommand\icbxcam[1]{\includegraphics[width=.08\textwidth]{images/balanced/xgrad-cam/test-pr=-1/#1} & \includegraphics[width=.08\textwidth]{images/balanced/xgrad-cam/test-pr=1/#1}}
\newcommand\icuxcam[1]{\includegraphics[width=.08\textwidth]{images/unbalanced/xgrad-cam/test-pr=-1/#1} & \includegraphics[width=.08\textwidth]{images/unbalanced/xgrad-cam/test-pr=1/#1}}
\newcommand\headhires[2]{\multicolumn{2}{c}{\tiny\bf{#1} -- #2\,\%}}
\newcommand\icbhires[1]{\includegraphics[width=.08\textwidth]{images/balanced/hirescam/test-pr=-1/#1} & \includegraphics[width=.08\textwidth]{images/balanced/hirescam/test-pr=1/#1}}
\newcommand\icuhires[1]{\includegraphics[width=.08\textwidth]{images/unbalanced/hirescam/test-pr=-1/#1} & \includegraphics[width=.08\textwidth]{images/unbalanced/hirescam/test-pr=1/#1}}
\newcommand\headgcamplus[2]{\multicolumn{2}{c}{\tiny\bf{#1} -- #2\,\%}}
\newcommand\icbgcamplus[1]{\includegraphics[width=.08\textwidth]{images/balanced/grad-cam++/test-pr=-1/#1} & \includegraphics[width=.08\textwidth]{images/balanced/grad-cam++/test-pr=1/#1}}
\newcommand\icugcamplus[1]{\includegraphics[width=.08\textwidth]{images/unbalanced/grad-cam++/test-pr=-1/#1} & \includegraphics[width=.08\textwidth]{images/unbalanced/grad-cam++/test-pr=1/#1}}
\newcommand\headc[2]{\multicolumn{2}{c}{\tiny\bf{#1} -- #2}}

\begin{table*}[t]
  \Caption[tab:error-rates]{Error Rates and Proportional Energy}{This table shows the probability of the positive class $p_m$, the False Negative and False Positive Rates, as well as Proportional Energy of Grad-CAM visualizations for positively and negatively predicted samples for the different attributes when extracted with an unbalanced and a balanced classifier. The attributes are sorted by increasing imbalance (deviation of $p_m$ from 0.5). For AFFACT-u, the classification error of the minority class, and the proportional energy of the majority class are bolded to highlight the problems arising for the unbalanced classifier.}\vspace*{-1ex}
  \centering\small
  \begin{tabularx}{.99\textwidth}{lc|CC|CC||CC|CC}
    && \multicolumn{4}{c||}{\bf Error Rates $\downarrow$} & \multicolumn{4}{c}{\bf Proportional Energy $\uparrow$}\\[.5ex]
    && \multicolumn{2}{c|}{\bf AFFACT-u} & \multicolumn{2}{c||}{\bf AFFACT-b} & \multicolumn{2}{c|}{\bf AFFACT-u} & \multicolumn{2}{c}{\bf AFFACT-b}\\[.5ex]
    \bf Attribute & \bf $p_m$ & \bf FNR & \bf FPR & \bf FNR & \bf FPR & \bf Pos & \bf Neg & \bf Pos & \bf Neg\\\hline
     Attractive          & 0.514 & 0.175     & \bf 0.174 & 0.169 & 0.177 & \bf 0.951 & 0.591     & 0.892 & 0.708 \\
 Mouth Slightly Open & 0.482 & \bf 0.055 & 0.048     & 0.053 & 0.048 & 0.563     & \bf 0.637 & 0.565 & 0.641 \\
 Smiling             & 0.480 & \bf 0.072 & 0.063     & 0.065 & 0.064 & 0.556     & \bf 0.547 & 0.554 & 0.651 \\
 Wearing Lipstick    & 0.470 & \bf 0.065 & 0.037     & 0.054 & 0.043 & 0.406     & \bf 0.374 & 0.458 & 0.446 \\
 High Cheekbones     & 0.452 & \bf 0.126 & 0.102     & 0.132 & 0.104 & 0.713     & \bf 0.603 & 0.672 & 0.736 \\
 Male                & 0.419 & \bf 0.017 & 0.008     & 0.017 & 0.008 & 0.986     & \bf 0.889 & 0.997 & 0.981 \\
 Heavy Makeup        & 0.384 & \bf 0.113 & 0.046     & 0.065 & 0.081 & 0.619     & \bf 0.242 & 0.683 & 0.590 \\
 Wavy Hair           & 0.319 & \bf 0.258 & 0.058     & 0.187 & 0.093 & 0.470     & \bf 0.096 & 0.472 & 0.201 \\
 Oval Face           & 0.283 & \bf 0.598 & 0.077     & 0.286 & 0.319 & 0.733     & \bf 0.168 & 0.528 & 0.484 \\
 Pointy Nose         & 0.276 & \bf 0.629 & 0.055     & 0.336 & 0.209 & 0.543     & \bf 0.253 & 0.520 & 0.447 \\
 Arched Eyebrows     & 0.266 & \bf 0.274 & 0.110     & 0.119 & 0.221 & 0.621     & \bf 0.064 & 0.686 & 0.309 \\
 Big Lips            & 0.241 & \bf 0.651 & 0.076     & 0.353 & 0.269 & 0.298     & \bf 0.079 & 0.259 & 0.176 \\
 Black Hair          & 0.239 & \bf 0.202 & 0.051     & 0.086 & 0.139 & 0.408     & \bf 0.092 & 0.390 & 0.326 \\
 Big Nose            & 0.236 & \bf 0.367 & 0.115     & 0.165 & 0.266 & 0.621     & \bf 0.113 & 0.596 & 0.417 \\
 Young               & 0.779 & 0.055     & \bf 0.300 & 0.137 & 0.151 & \bf 0.330 & 0.962     & 0.734 & 0.900 \\
 Straight Hair       & 0.209 & \bf 0.425 & 0.074     & 0.134 & 0.232 & 0.381     & \bf 0.311 & 0.397 & 0.439 \\
 Bags Under Eyes     & 0.204 & \bf 0.362 & 0.100     & 0.141 & 0.221 & 0.485     & \bf 0.093 & 0.416 & 0.488 \\
 Brown Hair          & 0.204 & \bf 0.248 & 0.074     & 0.115 & 0.205 & 0.472     & \bf 0.112 & 0.441 & 0.339 \\
 Wearing Earrings    & 0.187 & \bf 0.181 & 0.061     & 0.073 & 0.138 & 0.701     & \bf 0.126 & 0.686 & 0.403 \\
 No Beard            & 0.834 & 0.024     & \bf 0.081 & 0.044 & 0.036 & \bf 0.181 & 0.527     & 0.732 & 0.527 \\
 Bangs               & 0.152 & \bf 0.119 & 0.020     & 0.038 & 0.053 & 0.764     & \bf 0.013 & 0.704 & 0.729 \\
 Blond Hair          & 0.149 & \bf 0.153 & 0.017     & 0.048 & 0.067 & 0.332     & \bf 0.018 & 0.298 & 0.223 \\
 Bushy Eyebrows      & 0.144 & \bf 0.326 & 0.031     & 0.147 & 0.110 & 0.691     & \bf 0.005 & 0.641 & 0.330 \\
 Wearing Necklace    & 0.121 & \bf 0.507 & 0.038     & 0.152 & 0.230 & 0.633     & \bf 0.020 & 0.580 & 0.302 \\
 Narrow Eyes         & 0.116 & \bf 0.679 & 0.021     & 0.252 & 0.172 & 0.522     & \bf 0.023 & 0.514 & 0.666 \\
 5 o'Clock Shadow    & 0.112 & \bf 0.216 & 0.034     & 0.050 & 0.110 & 0.533     & \bf 0.033 & 0.482 & 0.572 \\
 Receding Hairline   & 0.080 & \bf 0.398 & 0.026     & 0.092 & 0.129 & 0.716     & \bf 0.004 & 0.724 & 0.528 \\
 Wearing Necktie     & 0.073 & \bf 0.159 & 0.014     & 0.036 & 0.060 & 0.805     & \bf 0.002 & 0.791 & 0.215 \\
 Rosy Cheeks         & 0.065 & \bf 0.361 & 0.026     & 0.044 & 0.144 & 0.563     & \bf 0.002 & 0.571 & 0.405 \\
 Eyeglasses          & 0.065 & \bf 0.018 & 0.002     & 0.013 & 0.007 & 0.710     & \bf 0.001 & 0.770 & 0.571 \\
 Goatee              & 0.064 & \bf 0.170 & 0.018     & 0.007 & 0.069 & 0.374     & \bf 0.004 & 0.413 & 0.420 \\
 Chubby              & 0.058 & \bf 0.420 & 0.024     & 0.073 & 0.154 & 0.981     & \bf 0.030 & 0.964 & 0.679 \\
 Sideburns           & 0.056 & \bf 0.116 & 0.019     & 0.014 & 0.070 & 0.539     & \bf 0.005 & 0.472 & 0.285 \\
 Blurry              & 0.051 & \bf 0.476 & 0.010     & 0.061 & 0.110 & 0.914     & \bf 0.041 & 0.843 & 0.726 \\
 Wearing Hat         & 0.049 & \bf 0.078 & 0.004     & 0.028 & 0.015 & 0.886     & \bf 0.002 & 0.872 & 0.622 \\
 Double Chin         & 0.047 & \bf 0.477 & 0.014     & 0.060 & 0.142 & 0.187     & \bf 0.001 & 0.254 & 0.045 \\
 Pale Skin           & 0.043 & \bf 0.491 & 0.007     & 0.048 & 0.147 & 0.801     & \bf 0.003 & 0.742 & 0.726 \\
 Gray Hair           & 0.042 & \bf 0.217 & 0.009     & 0.025 & 0.061 & 0.290     & \bf 0.009 & 0.297 & 0.103 \\
 Mustache            & 0.041 & \bf 0.505 & 0.011     & 0.034 & 0.081 & 0.421     & \bf 0.004 & 0.512 & 0.573 \\
 Bald                & 0.023 & \bf 0.191 & 0.005     & 0.025 & 0.032 & 0.744     & \bf 0.001 & 0.687 & 0.431 \\

  \end{tabularx}
\end{table*}

To verify the expected behavior of our attribute classifiers, we first compute the classification results for the different attributes on our selected frontal images from the test set.
The classification results, separated into False Negative Rate, \ie, the number of positively labeled samples wrongly predicted as negative, and False Positive Rate, can be found in \tabref{tab:error-rates}.
For easier access, we ordered the attributes by imbalance, starting with the almost balanced attributes, and ending with highly imbalanced ones.
The unbalanced network AFFACT-u works well on both classes as long as the attributes are rather balanced, but already at a small negative/positive imbalance of 68\%/32\% in the \texttt{Wavy Hair} attribute, the prediction of the majority class is about one magnitude better than that of the minority class.
For slightly less balanced attributes with about 75\%/25\% distribution, \texttt{Oval Face, Pointy Nose} and \texttt{Big Lips}, the minority class is even below random performance while the majority class enjoys reasonable classification accuracy.

When using our balancing technique to arrive at the AFFACT-b network, we can observe that False Negative Rates and False Positive Rates are distributed more evenly.
Hence, AFFACT-b is able to classify minority and majority classes similarly well, and no error rate goes beyond random chance.
However, we observe smaller FNRs than FPRs for many attributes, which might indicate that the presence of a facial attribute is easier to classify than its absence --- or that the presence of the attribute is more consistently labeled than its absence, cf.~\cite{wu2023consistency} for a more detailed label analysis.

\subsection{Visualization}
\label{sec:visualization}

\begin{figure*} [t]
    \centering
    \setlength{\tabcolsep}{1.5pt}
    \renewcommand{\arraystretch}{0.5}
    \begin{tabular}{*{4}{c@{\,}c@{\hspace*{1em}}}c@{\,}c}
    \head{Attractive}{51} & \head{Smiling}{48} & \head{Wearing Lipstick}{47} & \head{High Cheekbones}{45} & \head{Heavy Makeup}{38}\\
    \icu{Attractive} & \icu{Smiling} & \icu{Wearing_Lipstick} & \icu{High_Cheekbones} & \icu{Heavy_Makeup} \\
    \icb{Attractive} & \icb{Smiling} & \icb{Wearing_Lipstick} & \icb{High_Cheekbones} & \icb{Heavy_Makeup} \\[.5ex]

    \head{Oval Face}{28} & \head{Arched Eyebrows}{27} & \head{Brown Hair}{20} & \head{Wearing Earrings}{19} & \head{Bushy Eyebrows}{14}\\
    \icu{Oval_Face} & \icu{Arched_Eyebrows} & \icu{Brown_Hair} & \icu{Wearing_Earrings} & \icu{Bushy_Eyebrows} \\
    \icb{Oval_Face} & \icb{Arched_Eyebrows} & \icb{Brown_Hair} & \icb{Wearing_Earrings} & \icb{Bushy_Eyebrows} \\[.5ex]

    \head{Wearing Necktie}{12} & \head{5 o'Clock Shadow}{11} & \head{Rosy Cheeks}{6} & \head{Blurry}{5} & \head{Bald}{2}\\
    \icu{Wearing_Necktie} & \icu{5_o_Clock_Shadow} & \icu{Rosy_Cheeks} & \icu{Blurry} & \icu{Brown_Hair} \\
    \icb{Wearing_Necktie} & \icb{5_o_Clock_Shadow} & \icb{Rosy_Cheeks} & \icb{Blurry} & \icb{Brown_Hair}

    \end{tabular}

    \Caption[fig:grad-cam-activation]{Averaged Grad-CAM Activations}{This figure displays the average CAM activations for 15 different attributes including the probability of positive label $p_m$. Activations are averaged across all negative (left) and positive (right) predictions, extracted by AFFACT-u (top) and AFFACT-b (bottom).}
\end{figure*}

Having observed that AFFACT-u classifies majority classes well, we expect that this decision is based on reasonable features from the images, while minority class samples with much worse classification performance rely on more dubious features.
In \figref{fig:grad-cam-activation} we can observe the average activation of our inputs via Grad-CAM, where each pair of images includes the average of all negative predicted attributes on the left, and the average of positive predictions on the right.
Again, the attributes are ordered by increasing imbalance, and the visualizations of all attributes can be found in the supplemental.
For the most balanced attributes in the first four results from the top row of \figref{fig:grad-cam-activation}, we can see that AFFACT-u generally makes use of reasonable features, for both classes, and the classification of the presence of an attribute has a larger activated region in the image.

Starting already in the second row and continuing to the bottom, the visualization of the \emph{majority} class tends to rely on the bottom-left corner (sometimes also the other corners) of the image, like \texttt{Rosy Cheeks} in \figref{fig:grad-cam-activation}, or do not show any activation whatsoever like \texttt{Bushy Eyebrows} and \texttt{Blurry}.
The latter can be explained by the fact that the prediction of the majority class solely relies on the bias neuron $b$ in \eqref{eq:logit} and is not influenced by any feature extracted from the image.
When a corner shows activation, we interpret that the network has learned that no relevant features can be extracted from the corners, so these are activated independently of the image input, and they serve as another bias unit similar to $b$.
Now, since the network has learned to use the corners as bias units, also more balanced attributes, such as \texttt{Attractive, Wearing Lipstick} or \texttt{High Cheekbones} can assign some energy to these locations that would otherwise be assigned to the bias neuron $b$.

When looking into the visualizations for the AFFACT-b network in \figref{fig:grad-cam-activation}, one can observe that valid features are extracted for both classes in each attribute, although the minority prediction generally has larger activated regions.
This effect can be explained by assuming that minority classes extract stronger features since these samples are weighted higher during training, but it can also be an effect of the dataset where most of the minority classes represent the presence of attributes, and the presence generally can rely on a larger set of features then predicting absence of attributes.
Anyway, in no case there is any activation in the corners of the images, so the network has successfully learned to ignore the corners that do not include useful information for the classification.

\begin{table*}[t]
  \Caption[tab:dif-cams]{Proportional Energy for Different Techniques}{This table shows the Proportional Energy for positively and negatively predicted samples for the different attributes when extracted with an unbalanced and a balanced classifier. The highest proportional energy of the majority class are bolded.}\vspace*{-1ex}
  \centering\small
  \begin{tabularx}{.9\textwidth}{l|c||CC|CC}
    && \multicolumn{2}{c|}{\bf AFFACT-u} & \multicolumn{2}{c}{\bf AFFACT-b} \\[.5ex]
    \bf Attribute & \bf Method & \bf Pos & \bf Neg& \bf Pos & \bf Neg \\\hline
                        & GradCAM               & 0.951     & 0.591     & 0.892     & 0.728\\
Attractive          & Grad-CAM++            & 0.063     & 0.214     & 0.049     & 0.121\\
($p_m=0.514$)       & HiResCAM              & 0.977     & 0.784     & 0.966     & 0.916\\
                    & Element-wise Grad-CAM    & 0.954     & \bf 0.889 & 0.948     & \bf 0.940\\ \hline

                    & GradCAM               & 0.701     & 0.126     & 0.686     & 0.403\\
Wearing Earrings    & Grad-CAM++            & 0.840     & 0.075     & 0.578     & 0.116\\
($p_m=0.187$)       & HiResCAM              & 0.756     & 0.294     & 0.647     & \bf 0.445\\
                    & Element-wise Grad-CAM    & 0.631     & \bf 0.326 & 0.504     & 0.401\\ \hline

                    & GradCAM               & 0.744     & 0.001     & 0.687     & \bf 0.431\\
Bald                & Grad-CAM++            & 0.824     & 0.000     & 0.835     & 0.108\\
($p_m=0.023$)       & HiResCAM              & 0.784     & 0.141     & 0.752     & 0.370\\
                    & Element-wise Grad-CAM   & 0.699     & \bf 0.158 & 0.540     & 0.311\\

  \end{tabularx}
\end{table*}

\subsection{Proportional Energy}

To provide a numerical evaluation of the visualizations, we make use of the proportional energy \eqref{eq:prop-energy} that we compute using the masks defined in the supplemental material and in \cite{bieri2023bachelor}.
Since the size of the masks differs between attributes, the absolute values of proportional energy cannot be compared across attributes.
Again, we split the results into samples predicted as positive and as negative, and compare the unbalanced and the balanced network.
The average proportional energy over all respective samples and for all attributes can be found in \tabref{tab:error-rates}.
With these results, we can numerically verify the trend that we could also observe in \figref{fig:grad-cam-activation}.
As soon as the imbalance crosses the 40\%/60\% border, \ie, starting from \texttt{Heavy Makeup} the proportional energy for predicting the majority class via AFFACT-u reduces dramatically when compared to predicting the minority class, which proves strongly that AFFACT-u needs to depend on more dubious features to predict majority classes.
For AFFACT-b, there also exist differences in the prediction of presence or absence of features, but these are rarely as pronounced as for AFFACT-u.

Most of these differences highlight that predicting the presence of an attribute might be more localized than predicting its absence, which might include other parts of the face as well.
For example, the prediction of the presence of \texttt{Wearing Earrings} or \texttt{Wearing Necklace} need to focus more closely to the ear or neck region, while the absence of these can also include locations that indicate the gender of the person --- since the presence of such attributes correlate with gender, which can better be approximated from the full face, see the highlighted locations in \figref{fig:grad-cam-activation}.

\subsection{CAM Techniques}
\begin{figure*} [t]
    \centering
    \setlength{\tabcolsep}{1.5pt}
    \renewcommand{\arraystretch}{0.5}

    \begin{tabular}{*{3}{c@{\,}c@{\hspace*{1em}}}c@{\ }c}

    \headc{Attractive}{GC}  & \headc{Attractive}{GC++} & \headc{Attractive}{HR} & \headc{Attractive}{EW}\\
    \icu{Attractive} & \icugcamplus{Attractive} & \icuhires{Attractive} & \icuelementwise{Attractive}\\
    \icb{Attractive} & \icbgcamplus{Attractive} & \icbhires{Attractive} & \icbelementwise{Attractive}\\[.5ex]

    \headc{Wearing Earrings}{GC} &\headc{Wearing Earrings}{GC++} & \headc{Wearing Earrings}{HR}& \headc{Wearing Earrings}{EW}\\
    \icu{Wearing_Earrings} & \icugcamplus{Wearing_Earrings} & \icuhires{Wearing_Earrings} & \icuelementwise{Wearing_Earrings}\\
    \icb{Wearing_Earrings} & \icbgcamplus{Wearing_Earrings} & \icbhires{Wearing_Earrings} & \icbelementwise{Wearing_Earrings}\\[.5ex]

    \headc{Bald}{GC} &\headc{Bald}{GC++} & \headc{Bald}{HR}& \headc{Bald}{EW}\\
    \icu{Bald} & \icugcamplus{Bald} & \icuhires{Bald} & \icuelementwise{Bald}\\
    \icb{Bald} & \icbgcamplus{Bald} & \icbhires{Bald} & \icbelementwise{Bald}

    \end{tabular}

    \Caption[fig:combined]{Averaged Activations for Different CAM Techniques}{This figure shows the average activations for Grad-CAM (GC), GradCAM++ (GC++), HiResCAM (HR) and element-wise CAM (EW). Blocks are built identically to \figref{fig:grad-cam-activation}.}
\end{figure*}

When comparing the visualizations of the same attribute from different CAM techniques, we observe that HiResCAM and Element-wise Grad-CAM can visualize the feature maps better for models with negative labels in \figref{fig:combined} and the supplemental, especially for AFFACT-u.
At the same time, we realize that the results of proportional energy for HiResCAM and Element-wise Grad-CAM are obviously larger than others in \tabref{tab:dif-cams}.
It indicates that these two techniques can produce visualization results that are clearer and more comprehensive.
While HiResCAM and Element-wise CAM include some more information in the prediction of majority classes in AFFACT-u, the corners of the images are still often activated.
Additionally, the mouth region is activated for majority classes, independent of the attribute (best seen in the supplemental).
GradCAM++ does not extract information of majority classes even for AFFACT-b, we believe that this is an artifact of the visualization technique.

\subsection{Target Classes}

One of the contributions of this paper is the extension of the categorical classifier visualization to binary classifiers that only have one output node.
Here, we show the impact of the visualization when targeting the positive class only, which is what is done in categorical classifiers.
Particularly, we use the weight $\alpha_k$ in \eqref{eq:gradcam-binary} without computing the absolute value:
\begin{equation}
  \label{eq:gradcam-categorical}
  \alpha_k = \sum_{(i,j)} \frac{\partial z}{\partial f_k(i,j)}
\end{equation}
Notably, for positively predicted classes, \ie~where $z>0$, both \eqref{eq:gradcam-binary} and \eqref{eq:gradcam-categorical} result in the same visualization.
Hence, in \figref{fig:targetclass} we show the impact of our proposed method on negatively-predicted samples for different attributes with various imbalance.
To avoid influences of unbalanced predictions discussed in \secref{sec:visualization}, we utilize our balanced network AFFACT-b in these visualizations.
As can be clearly seen in \figref{fig:targetclass}, the positive class visualization on the left side of each pair highlights various different regions but the ones that would be expected.
Only the predicted class visualization on the right concentrates on the correct part of the image.

\section{Conclusion}

In our work, we have modified gradient-based class activation mapping techniques to work with single-output binary classifiers and then applied them to facial attribute classification.
We investigated an unbalanced classifier AFFACT-u and showed that this classifier produces extremely low classification errors on majority classes.
The visualization results from different CAM techniques prove that these decisions are almost solely based on the bias neuron of the final classification layer or some corners of the image, but not on reasonable areas of the images.
On the other hand, minority classes are predicted with extremely high error, sometimes beyond random guessing, but the visualization highlights reasonable regions in the image.
Due to the nature of training the classification of several attributes jointly, negative effects from highly imbalanced attributes, \ie, classifying from the corners of the image, are transferred to more balanced attributes.

When applying a training scheme that balances the imbalanced classes, we arrived at the AFFACT-b model, which showed much more reasonable behavior, both in the classification of minority classes (on the cost of misclassifying majority class samples more often), and in the visualization of input regions.
While the visualizations in \figref{fig:grad-cam-activation} and \figref{fig:combined} show only averages, we still observed a few cases where even the balanced classifier has no active regions in the image, especially in highly imbalanced classes, so further research has to be done to understand this corner-case behavior.

\newcommand\icc[1]{\includegraphics[width=.077\textwidth]{images/comparison/class_target/#1} & \includegraphics[width=.077\textwidth]{images/comparison/prediction_target/#1}}
\begin{figure*}[tb]
  \centering
  \subfloat[Not \texttt{Smiling}]{
    \begin{tabular}{c@{\ }c}
      \icc{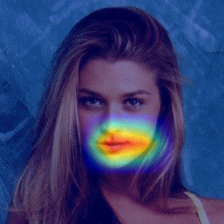}\\
      \icc{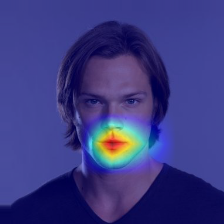}
    \end{tabular}
  }
  \subfloat[Not \texttt{Male}]{
    \begin{tabular}{c@{\ }c}
      \icc{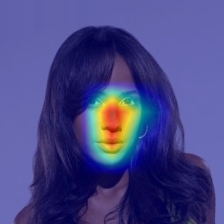}\\
      \icc{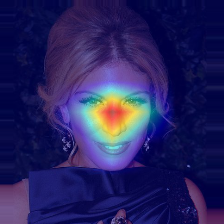}
    \end{tabular}
  }
  \subfloat[Not \texttt{Double Chin}]{
    \begin{tabular}{c@{\ }c}
      \icc{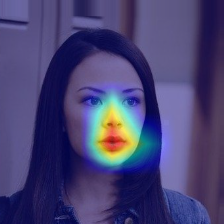}\\
      \icc{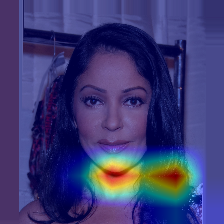}
    \end{tabular}
  }
  \subfloat[Not \texttt{Eyeglasses}]{
    \begin{tabular}{c@{\ }c}
      \icc{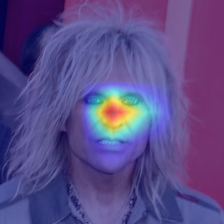}\\
      \icc{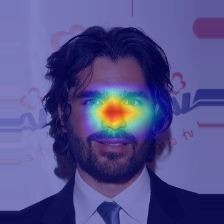}
    \end{tabular}
  }
  \Caption[fig:targetclass]{AFFACT-b Negative Class Visualization}{This figure shows Grad-CAM visualizations of samples for four different attributes that were negatively predicted by the balanced network. On the left of each block, we visualize the categorical target via \eqref{eq:gradcam-categorical}, \ie, the positive class. On the right we present the visualization of the predicted negative class created with \eqref{eq:gradcam-binary}.}

\end{figure*}

\subsection{Discussion}
In this work, we have only used two binary attribute prediction networks with the same network topology, and further studies would be required to validate our findings on other binary classification tasks and other network topologies.
Additionally, we just have made use of the Euclidean loss function, but initial experiments indicate that our findings translate to binary cross-entropy loss and binary classification networks with two outputs trained with SoftMax loss, though activation patterns seem to differ slightly.
Also, we have used class weights to provide a balanced network, the influence of FocalLoss \citep{lin2017focalloss} or other approaches for balancing classes \cite{menon2020long,rangwani2022escaping} would need further investigation.
Besides, we have applied a variety of CAM approaches like GradCAM, Grad-CAM++, HiResCAM, and Element-wise Grad-CAM in our experiments, which provided slightly different views on our conclusion.
We also planned to show results for the FullGrad method \citep{srinivas2019fullgrad}, but the available implementation of \cite{gildenblat2021pytorchcam} was too slow to run on the large-scale dataset in reasonable time.
Finally, our implementation of the binary classifier target only applies to gradient-based CAM techniques, the extension to non gradient-based techniques such as ScoreCAM \citep{wang2020score-cam} remains unsolved for now.
Also, the visualization of some gradient-based methods such as XGradCAM \citep{fu2020xgradcam} do not work with our extension, which needs to be investigated.

For the computation of proportional energy, we have defined some masks that contain reasonable regions in the images.
While we have taken care that the masks cover all parts of the image that we deem useful for the prediction of the presence of that attribute, a better definition of masks will improve proportional energy values for some attributes.
However, the overall conclusion in our paper will likely not be influenced by better masks.

\iftrue
\section*{Acknowledgements}
{\fontsize{8}{8}\selectfont\noindent
This research is based upon work supported in part by the Office of the Director of National Intelligence (ODNI), Intelligence Advanced Research Projects Activity (IARPA), via 2022-21102100003.
The views and conclusions contained herein are those of the authors and should not be interpreted as necessarily representing the official policies,
either expressed or implied, of ODNI, IARPA, or the U.S.~Government.
The U.S.~Government is authorized to reproduce and distribute reprints for governmental purposes notwithstanding any copyright annotation therein.}\vspace*{-1ex}
\fi
{\small
\bibliography{texfiles/Publications,texfiles/References}
\bibliographystyle{ieeetr}}

\iftrue
\newpage
\begin{onecolumn}
\section{Supplemental}

\subsection{Attribute Masks}
Attribute masks were generated based on the fact that most CAM methods work on the final convolutional layer.
Since we utilize ResNet-50 as basis, the final convolutional layer reduces the original input image size of $224\times224$ pixels to a feature map of resolution $7\times7$, each cell of the feature map represents $32\times32$ pixels of the input.
Thus, we define our masks in terms of $32\times32$ blocks, based on the intuition onto which parts of the face the classifier should base its decision on.
You can find the masks for the different attributes in \figref{fig:masks}, some attributes looking into similar regions share the masks.
Slightly different masks would likely improve the proportional energy calculation for various attributes, but are out of scope of this work.
Notably, none of our masks ever include any corner of the image.

\newcommand\img[1]{\includegraphics[width=.2\textwidth]{images/masks/#1_overlay}}
\begin{figure*}[hb]
  \centering
  \subfloat[\label{fig:masks:masks}Masks]{
  \begin{tabularx}{.95\textwidth}{CCCC}
    Head & Hair & Eyes & Face \\
    \img{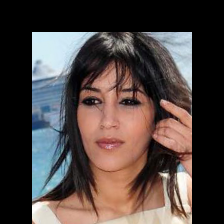} & \img{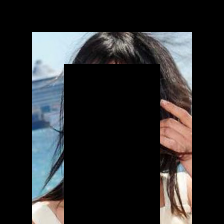} & \img{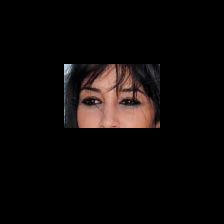} & \img{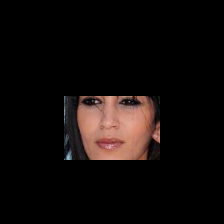} \\
    Lips & Ears & Neck & Forehead\\
    \img{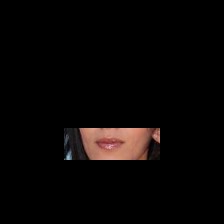} & \img{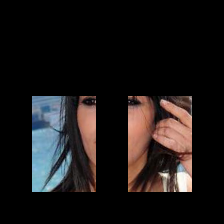} & \img{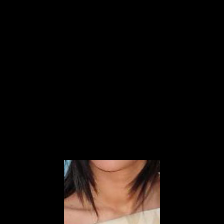} & \img{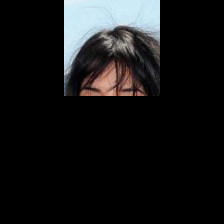}\\
    Shape & \texttt{Double Chin} & \texttt{Goatee} & \texttt{Wearing Hat}  \\
    \img{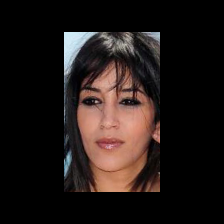} & \img{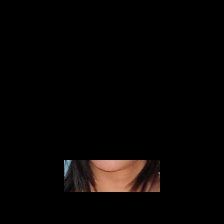} & \img{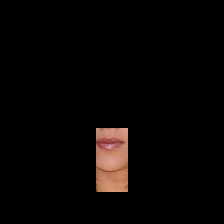} & \img{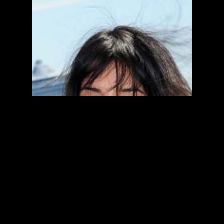}
  \end{tabularx}}

  \subfloat[\label{fig:masks:attributes}Attributes]{
  \begin{tabularx}{.85\textwidth}{c|X}
    Head & \texttt{Attractive, Blurry, Chubby, Male, Young} \\
    Hair & \texttt{Black Hair, Blonde Hair, Brown Hair, Gray Hair, Straight Hair, Wavy Hair}\\
    Eyes & \texttt{Arched Eyebrows, Bags Under Eyes, Bangs, Bushy Eyebrows, Eyeglasses, Narrow Eyes}\\
    Face & \texttt{5 o'Clock Shadow, Big Nose, Heavy Makeup, High Cheekbones, No Beard, Pointy Nose, Rosy Cheeks}\\
    Lips & \texttt{Big Lips, Mouth Slightly Open, Mustache, Smiling, Wearing Lipstick}\\
    Ears & \texttt{Sideburns, Wearing Earrings} \\
    Neck & \texttt{Wearing Necklace, Wearing Necktie} \\
    Forehead & \texttt{Bald, Receding Hairline} \\
    Shape & \texttt{Oval Face, Pale Skin}
  \end{tabularx}}

  \Caption[fig:masks]{Attribute Masks}{The images in \subref*{fig:masks:masks} show the different defined masks, applied to one input image. \subref*{fig:masks:attributes} lists the attributes for which the masks are valid for. The last three masks are defined for single attributes.}
\end{figure*}

\subsection{Proportional Energy and Visualization}
In the main paper, we had only listed results of a few attributes.
The remaining attributes, which we sort by increasing imbalance, can be found here.
In \tabref{tab:result-all}, we show the proportional energy of AFFACT-u and AFFACT-b for four CAM techniques, averaged for positively and negatively predicted attributes.
In Figures~\ref{fig:grad-cam}-\ref{fig:elementwisecam-activation}, you can find the average activations for these visualizations techniques.

\clearpage
\onecolumn
\begin{longtable}{l|c|c||cc|cc}
  \caption{\label{tab:result-all}\texttt{Proportional Energy.} \textit{This table includes Proportional Energy values obtained through four different CAM techniques, averaged for all positively and negatively predicted samples per attribute by two networks.}}\\
  \multicolumn{3}{c||}{} & \multicolumn{2}{c|}{\bf AFFACT-u} & \multicolumn{2}{c}{\bf AFFACT-b} \\
  \bf Attribute & $p_m$ & \bf Method & \bf Pos & \bf Neg & \bf Pos & \bf Neg \\
  \hline
  \endfirsthead
  \caption{(Continued)} \\
  \multicolumn{3}{c||}{} & \multicolumn{2}{c|}{\bf AFFACT-u} & \multicolumn{2}{c}{\bf AFFACT-b} \\
  \bf Attribute & $p_m$ & \bf Method & \bf Pos & \bf Neg & \bf Pos & \bf Neg \\
  \hline
  \endhead
  \hline
  \multicolumn{7}{r}{Continued on the next page} \\
  \endfoot
  \hline
  \endlastfoot

  \multirow{4}{*}{\small\tt Attractive} & \multirow{4}{*}{0.514}
      & GradCAM & 0.951 & 0.591 & 0.892 & 0.708 \\*
     && Grad-CAM++ & 0.063 & 0.214 & 0.049 & 0.121 \\*
     && HiResCAM & 0.977 & 0.784 & 0.966 & 0.916 \\*
     && Element-wise CAM & 0.954 & 0.889 & 0.948 & 0.940 \\\hline
\multirow{4}{*}{\small\tt Mouth Sl. Open} & \multirow{4}{*}{0.482}
      & GradCAM & 0.563 & 0.637 & 0.565 & 0.641 \\*
     && Grad-CAM++ & 0.647 & 0.715 & 0.648 & 0.705 \\*
     && HiResCAM & 0.597 & 0.412 & 0.563 & 0.489 \\*
     && Element-wise CAM & 0.501 & 0.393 & 0.456 & 0.409 \\\hline
\multirow{4}{*}{\small\tt Smiling} & \multirow{4}{*}{0.480}
      & GradCAM & 0.556 & 0.547 & 0.554 & 0.651 \\*
     && Grad-CAM++ & 0.702 & 0.669 & 0.685 & 0.695 \\*
     && HiResCAM & 0.614 & 0.342 & 0.540 & 0.481 \\*
     && Element-wise CAM & 0.500 & 0.354 & 0.429 & 0.401 \\\hline
\multirow{4}{*}{\small\tt Wearing Lipstick} & \multirow{4}{*}{0.470}
      & GradCAM & 0.406 & 0.374 & 0.458 & 0.446 \\*
     && Grad-CAM++ & 0.447 & 0.265 & 0.404 & 0.141 \\*
     && HiResCAM & 0.350 & 0.272 & 0.337 & 0.303 \\*
     && Element-wise CAM & 0.324 & 0.276 & 0.294 & 0.273 \\\hline
\multirow{4}{*}{\small\tt High Cheekbones} & \multirow{4}{*}{0.452}
      & GradCAM & 0.713 & 0.603 & 0.672 & 0.736 \\*
     && Grad-CAM++ & 0.833 & 0.673 & 0.722 & 0.474 \\*
     && HiResCAM & 0.698 & 0.481 & 0.639 & 0.609 \\*
     && Element-wise CAM & 0.604 & 0.513 & 0.558 & 0.546 \\\hline
\multirow{4}{*}{\small\tt Male} & \multirow{4}{*}{0.419}
      & GradCAM & 0.986 & 0.889 & 0.997 & 0.981 \\*
     && Grad-CAM++ & 0.863 & 0.553 & 0.742 & 0.846 \\*
     && HiResCAM & 0.979 & 0.962 & 0.985 & 0.976 \\*
     && Element-wise CAM & 0.964 & 0.944 & 0.973 & 0.963 \\\hline
\multirow{4}{*}{\small\tt Heavy Makeup} & \multirow{4}{*}{0.384}
      & GradCAM & 0.619 & 0.242 & 0.683 & 0.590 \\*
     && Grad-CAM++ & 0.585 & 0.227 & 0.475 & 0.216 \\*
     && HiResCAM & 0.632 & 0.419 & 0.599 & 0.562 \\*
     && Element-wise CAM & 0.578 & 0.459 & 0.541 & 0.508 \\\hline
\multirow{4}{*}{\small\tt Wavy Hair} & \multirow{4}{*}{0.319}
      & GradCAM & 0.470 & 0.096 & 0.472 & 0.201 \\*
     && Grad-CAM++ & 0.505 & 0.015 & 0.502 & 0.063 \\*
     && HiResCAM & 0.474 & 0.187 & 0.417 & 0.248 \\*
     && Element-wise CAM & 0.387 & 0.204 & 0.319 & 0.240 \\\hline
\multirow{4}{*}{\small\tt Oval Face} & \multirow{4}{*}{0.283}
      & GradCAM & 0.733 & 0.168 & 0.528 & 0.484 \\*
     && Grad-CAM++ & 0.070 & 0.014 & 0.009 & 0.037 \\*
     && HiResCAM & 0.752 & 0.387 & 0.696 & 0.558 \\*
     && Element-wise CAM & 0.739 & 0.633 & 0.729 & 0.719 \\\hline
\multirow{4}{*}{\small\tt Pointy Nose} & \multirow{4}{*}{0.276}
      & GradCAM & 0.543 & 0.253 & 0.520 & 0.447 \\*
     && Grad-CAM++ & 0.571 & 0.021 & 0.056 & 0.110 \\*
     && HiResCAM & 0.704 & 0.337 & 0.530 & 0.472 \\*
     && Element-wise CAM & 0.576 & 0.419 & 0.476 & 0.464 \\\hline
\multirow{4}{*}{\small\tt Arched Eyebrows} & \multirow{4}{*}{0.266}
      & GradCAM & 0.621 & 0.064 & 0.686 & 0.309 \\*
     && Grad-CAM++ & 0.764 & 0.019 & 0.400 & 0.149 \\*
     && HiResCAM & 0.702 & 0.242 & 0.599 & 0.403 \\*
     && Element-wise CAM & 0.560 & 0.322 & 0.485 & 0.418 \\\hline
\multirow{4}{*}{\small\tt Big Lips} & \multirow{4}{*}{0.241}
      & GradCAM & 0.298 & 0.079 & 0.259 & 0.176 \\*
     && Grad-CAM++ & 0.200 & 0.008 & 0.012 & 0.007 \\*
     && HiResCAM & 0.359 & 0.128 & 0.265 & 0.171 \\*
     && Element-wise CAM & 0.284 & 0.202 & 0.233 & 0.208 \\\hline
\multirow{4}{*}{\small\tt Black Hair} & \multirow{4}{*}{0.239}
      & GradCAM & 0.408 & 0.092 & 0.390 & 0.326 \\*
     && Grad-CAM++ & 0.381 & 0.011 & 0.253 & 0.142 \\*
     && HiResCAM & 0.400 & 0.263 & 0.366 & 0.302 \\*
     && Element-wise CAM & 0.341 & 0.250 & 0.291 & 0.267 \\\hline
\multirow{4}{*}{\small\tt Big Nose} & \multirow{4}{*}{0.236}
      & GradCAM & 0.621 & 0.113 & 0.596 & 0.417 \\*
     && Grad-CAM++ & 0.647 & 0.013 & 0.060 & 0.036 \\*
     && HiResCAM & 0.674 & 0.265 & 0.574 & 0.414 \\*
     && Element-wise CAM & 0.562 & 0.381 & 0.491 & 0.442 \\\hline
\multirow{4}{*}{\small\tt Young} & \multirow{4}{*}{0.779}
      & GradCAM & 0.330 & 0.962 & 0.734 & 0.900 \\*
     && Grad-CAM++ & 0.027 & 0.698 & 0.033 & 0.097 \\*
     && HiResCAM & 0.811 & 0.988 & 0.947 & 0.947 \\*
     && Element-wise CAM & 0.875 & 0.972 & 0.940 & 0.951 \\\hline
\multirow{4}{*}{\small\tt Straight Hair} & \multirow{4}{*}{0.209}
      & GradCAM & 0.381 & 0.311 & 0.397 & 0.439 \\*
     && Grad-CAM++ & 0.384 & 0.144 & 0.185 & 0.309 \\*
     && HiResCAM & 0.378 & 0.266 & 0.357 & 0.343 \\*
     && Element-wise CAM & 0.321 & 0.255 & 0.279 & 0.278 \\\hline
\multirow{4}{*}{\small\tt Bags Under Eyes} & \multirow{4}{*}{0.204}
      & GradCAM & 0.485 & 0.093 & 0.416 & 0.488 \\*
     && Grad-CAM++ & 0.722 & 0.021 & 0.094 & 0.131 \\*
     && HiResCAM & 0.580 & 0.222 & 0.412 & 0.448 \\*
     && Element-wise CAM & 0.486 & 0.306 & 0.412 & 0.414 \\\hline
\multirow{4}{*}{\small\tt Brown Hair} & \multirow{4}{*}{0.204}
      & GradCAM & 0.472 & 0.112 & 0.441 & 0.339 \\*
     && Grad-CAM++ & 0.470 & 0.025 & 0.255 & 0.156 \\*
     && HiResCAM & 0.478 & 0.234 & 0.420 & 0.300 \\*
     && Element-wise CAM & 0.401 & 0.233 & 0.316 & 0.267 \\\hline
\multirow{4}{*}{\small\tt Wearing Earrings} & \multirow{4}{*}{0.187}
      & GradCAM & 0.701 & 0.126 & 0.686 & 0.403 \\*
     && Grad-CAM++ & 0.840 & 0.075 & 0.578 & 0.116 \\*
     && HiResCAM & 0.756 & 0.294 & 0.647 & 0.445 \\*
     && Element-wise CAM & 0.631 & 0.326 & 0.504 & 0.401 \\\hline
\multirow{4}{*}{\small\tt No Beard} & \multirow{4}{*}{0.834}
      & GradCAM & 0.181 & 0.527 & 0.732 & 0.527 \\*
     && Grad-CAM++ & 0.096 & 0.572 & 0.426 & 0.576 \\*
     && HiResCAM & 0.364 & 0.530 & 0.575 & 0.509 \\*
     && Element-wise CAM & 0.396 & 0.527 & 0.517 & 0.495 \\\hline
\multirow{4}{*}{\small\tt Bangs} & \multirow{4}{*}{0.152}
      & GradCAM & 0.764 & 0.013 & 0.704 & 0.729 \\*
     && Grad-CAM++ & 0.833 & 0.005 & 0.818 & 0.448 \\*
     && HiResCAM & 0.808 & 0.287 & 0.747 & 0.586 \\*
     && Element-wise CAM & 0.737 & 0.312 & 0.625 & 0.502 \\\hline
\multirow{4}{*}{\small\tt Blond Hair} & \multirow{4}{*}{0.149}
      & GradCAM & 0.332 & 0.018 & 0.298 & 0.223 \\*
     && Grad-CAM++ & 0.265 & 0.001 & 0.230 & 0.063 \\*
     && HiResCAM & 0.318 & 0.232 & 0.322 & 0.242 \\*
     && Element-wise CAM & 0.302 & 0.224 & 0.287 & 0.236 \\\hline
\multirow{4}{*}{\small\tt Bushy Eyebrows} & \multirow{4}{*}{0.144}
      & GradCAM & 0.691 & 0.005 & 0.641 & 0.330 \\*
     && Grad-CAM++ & 0.822 & 0.002 & 0.771 & 0.090 \\*
     && HiResCAM & 0.809 & 0.223 & 0.758 & 0.400 \\*
     && Element-wise CAM & 0.698 & 0.296 & 0.575 & 0.436 \\\hline
\multirow{4}{*}{\small\tt Wearing Necklace} & \multirow{4}{*}{0.121}
      & GradCAM & 0.633 & 0.020 & 0.580 & 0.302 \\*
     && Grad-CAM++ & 0.842 & 0.000 & 0.305 & 0.025 \\*
     && HiResCAM & 0.744 & 0.125 & 0.525 & 0.249 \\*
     && Element-wise CAM & 0.514 & 0.148 & 0.313 & 0.211 \\\hline
\multirow{4}{*}{\small\tt Narrow Eyes} & \multirow{4}{*}{0.116}
      & GradCAM & 0.522 & 0.023 & 0.514 & 0.666 \\*
     && Grad-CAM++ & 0.818 & 0.006 & 0.154 & 0.613 \\*
     && HiResCAM & 0.792 & 0.263 & 0.423 & 0.680 \\*
     && Element-wise CAM & 0.657 & 0.328 & 0.434 & 0.521 \\\hline
\multirow{4}{*}{\small\tt 5 o'Clock Shadow} & \multirow{4}{*}{0.112}
      & GradCAM & 0.533 & 0.033 & 0.482 & 0.572 \\*
     && Grad-CAM++ & 0.608 & 0.011 & 0.388 & 0.187 \\*
     && HiResCAM & 0.538 & 0.285 & 0.532 & 0.517 \\*
     && Element-wise CAM & 0.517 & 0.346 & 0.495 & 0.487 \\\hline
\multirow{4}{*}{\small\tt Receding Hairline} & \multirow{4}{*}{0.080}
      & GradCAM & 0.716 & 0.004 & 0.724 & 0.528 \\*
     && Grad-CAM++ & 0.843 & 0.001 & 0.797 & 0.078 \\*
     && HiResCAM & 0.810 & 0.157 & 0.727 & 0.335 \\*
     && Element-wise CAM & 0.665 & 0.189 & 0.506 & 0.307 \\\hline
\multirow{4}{*}{\small\tt Wearing Necktie} & \multirow{4}{*}{0.073}
      & GradCAM & 0.805 & 0.002 & 0.791 & 0.215 \\*
     && Grad-CAM++ & 0.886 & 0.000 & 0.892 & 0.137 \\*
     && HiResCAM & 0.863 & 0.130 & 0.810 & 0.232 \\*
     && Element-wise CAM & 0.727 & 0.142 & 0.542 & 0.206 \\\hline
\multirow{4}{*}{\small\tt Rosy Cheeks} & \multirow{4}{*}{0.065}
      & GradCAM & 0.563 & 0.002 & 0.571 & 0.405 \\*
     && Grad-CAM++ & 0.793 & 0.000 & 0.589 & 0.149 \\*
     && HiResCAM & 0.707 & 0.230 & 0.663 & 0.465 \\*
     && Element-wise CAM & 0.635 & 0.303 & 0.552 & 0.477 \\\hline
\multirow{4}{*}{\small\tt Eyeglasses} & \multirow{4}{*}{0.065}
      & GradCAM & 0.710 & 0.001 & 0.770 & 0.571 \\*
     && Grad-CAM++ & 0.823 & 0.000 & 0.820 & 0.276 \\*
     && HiResCAM & 0.816 & 0.201 & 0.768 & 0.475 \\*
     && Element-wise CAM & 0.750 & 0.221 & 0.630 & 0.442 \\\hline
\multirow{4}{*}{\small\tt Goatee} & \multirow{4}{*}{0.064}
      & GradCAM & 0.374 & 0.004 & 0.413 & 0.420 \\*
     && Grad-CAM++ & 0.495 & 0.008 & 0.564 & 0.207 \\*
     && HiResCAM & 0.444 & 0.116 & 0.439 & 0.244 \\*
     && Element-wise CAM & 0.380 & 0.135 & 0.333 & 0.216 \\\hline
\multirow{4}{*}{\small\tt Chubby} & \multirow{4}{*}{0.058}
      & GradCAM & 0.981 & 0.030 & 0.964 & 0.679 \\*
     && Grad-CAM++ & 0.999 & 0.004 & 0.210 & 0.039 \\*
     && HiResCAM & 0.993 & 0.688 & 0.967 & 0.951 \\*
     && Element-wise CAM & 0.986 & 0.775 & 0.964 & 0.946 \\\hline
\multirow{4}{*}{\small\tt Sideburns} & \multirow{4}{*}{0.056}
      & GradCAM & 0.539 & 0.005 & 0.472 & 0.285 \\*
     && Grad-CAM++ & 0.498 & 0.007 & 0.353 & 0.083 \\*
     && HiResCAM & 0.522 & 0.235 & 0.454 & 0.415 \\*
     && Element-wise CAM & 0.490 & 0.264 & 0.416 & 0.394 \\\hline
\multirow{4}{*}{\small\tt Blurry} & \multirow{4}{*}{0.051}
      & GradCAM & 0.914 & 0.041 & 0.843 & 0.726 \\*
     && Grad-CAM++ & 0.984 & 0.002 & 0.936 & 0.032 \\*
     && HiResCAM & 0.976 & 0.782 & 0.819 & 0.958 \\*
     && Element-wise CAM & 0.914 & 0.756 & 0.911 & 0.944 \\\hline
\multirow{4}{*}{\small\tt Wearing Hat} & \multirow{4}{*}{0.049}
      & GradCAM & 0.886 & 0.002 & 0.872 & 0.622 \\*
     && Grad-CAM++ & 0.897 & 0.003 & 0.881 & 0.186 \\*
     && HiResCAM & 0.875 & 0.215 & 0.840 & 0.422 \\*
     && Element-wise CAM & 0.796 & 0.223 & 0.634 & 0.375 \\\hline
\multirow{4}{*}{\small\tt Double Chin} & \multirow{4}{*}{0.047}
      & GradCAM & 0.187 & 0.001 & 0.254 & 0.045 \\*
     && Grad-CAM++ & 0.159 & 0.000 & 0.080 & 0.005 \\*
     && HiResCAM & 0.200 & 0.064 & 0.259 & 0.108 \\*
     && Element-wise CAM & 0.176 & 0.081 & 0.170 & 0.120 \\\hline
\multirow{4}{*}{\small\tt Pale Skin} & \multirow{4}{*}{0.043}
      & GradCAM & 0.801 & 0.003 & 0.742 & 0.726 \\*
     && Grad-CAM++ & 0.915 & 0.000 & 0.744 & 0.067 \\*
     && HiResCAM & 0.907 & 0.408 & 0.791 & 0.766 \\*
     && Element-wise CAM & 0.892 & 0.502 & 0.802 & 0.775 \\\hline
\multirow{4}{*}{\small\tt Gray Hair} & \multirow{4}{*}{0.042}
      & GradCAM & 0.290 & 0.009 & 0.297 & 0.103 \\*
     && Grad-CAM++ & 0.255 & 0.005 & 0.167 & 0.012 \\*
     && HiResCAM & 0.278 & 0.228 & 0.270 & 0.216 \\*
     && Element-wise CAM & 0.252 & 0.216 & 0.230 & 0.218 \\\hline
\multirow{4}{*}{\small\tt Mustache} & \multirow{4}{*}{0.041}
      & GradCAM & 0.421 & 0.004 & 0.512 & 0.573 \\*
     && Grad-CAM++ & 0.642 & 0.003 & 0.670 & 0.290 \\*
     && HiResCAM & 0.574 & 0.152 & 0.469 & 0.358 \\*
     && Element-wise CAM & 0.460 & 0.174 & 0.365 & 0.301 \\\hline
\multirow{4}{*}{\small\tt Bald} & \multirow{4}{*}{0.023}
      & GradCAM & 0.744 & 0.001 & 0.687 & 0.431 \\*
     && Grad-CAM++ & 0.824 & 0.000 & 0.835 & 0.108 \\*
     && HiResCAM & 0.784 & 0.141 & 0.752 & 0.370 \\*
     && Element-wise CAM & 0.699 & 0.158 & 0.540 & 0.311 \\\hline

\end{longtable}
\clearpage
\twocolumn

\begin{figure*} [p]
  \centering
  \setlength{\tabcolsep}{1.5pt}
  \renewcommand{\arraystretch}{0.5}
  \scalebox{.85}{
  \begin{tabular}{*{4}{c@{\,}c@{\hspace*{1em}}}c@{\,}c}
  \head{Attractive}{51} & \head{Mouth Slgt. Open}{48} & \head{Smiling}{48} & \head{Wearing Lipstick}{47} & \head{High Cheekbones}{45}\\
  \icu{Attractive} & \icu{Mouth_Slightly_Open} & \icu{Smiling} & \icu{Wearing_Lipstick} & \icu{High_Cheekbones} \\
  \icb{Attractive} & \icb{Mouth_Slightly_Open} & \icb{Smiling} & \icb{Wearing_Lipstick} & \icb{High_Cheekbones} \\[.5ex]

  \head{Male}{42} & \head{Heavy Makeup}{38} & \head{Wavy Hair}{32} & \head{Oval Face}{28} & \head{Pointy Nose}{28}\\
  \icu{Male} & \icu{Heavy_Makeup} & \icu{Wavy_Hair} & \icu{Oval_Face} & \icu{Pointy_Nose} \\
  \icb{Male} & \icb{Heavy_Makeup} & \icb{Wavy_Hair} & \icb{Oval_Face} & \icb{Pointy_Nose} \\[.5ex]

  \head{Arched Eyebrows}{27} & \head{Big Lips}{24} & \head{Black Hair}{24} & \head{Big Nose}{24} & \head{Young}{78}\\
  \icu{Arched_Eyebrows} & \icu{Big_Lips} & \icu{Black_Hair} & \icu{Big_Nose} & \icu{Young} \\
  \icb{Arched_Eyebrows} & \icb{Big_Lips} & \icb{Black_Hair} & \icb{Big_Nose} & \icb{Young} \\[.5ex]

  \head{Straight Hair}{21} & \head{Brown Hair}{20} & \head{Bags Under Eyes}{20} & \head{Wearing Earrings}{19} & \head{No Beard}{83}\\
  \icu{Straight_Hair} & \icu{Brown_Hair} & \icu{Bags_Under_Eyes} & \icu{Wearing_Earrings} & \icu{No_Beard} \\
  \icb{Straight_Hair} & \icb{Brown_Hair} & \icb{Bags_Under_Eyes} & \icb{Wearing_Earrings} & \icb{No_Beard} \\[.5ex]

  \head{Bangs}{15} & \head{Blond Hair}{15} & \head{Bushy Eyebrows}{14} & \head{Wearing Necklace}{12} & \head{Narrow Eyes}{12}\\
  \icu{Bangs} & \icu{Blond_Hair} & \icu{Bushy_Eyebrows} & \icu{Wearing_Necklace} & \icu{Narrow_Eyes} \\
  \icb{Bangs} & \icb{Blond_Hair} & \icb{Bushy_Eyebrows} & \icb{Wearing_Necklace} & \icb{Narrow_Eyes} \\[.5ex]

  \head{5 o'Clock Shadow}{11} & \head{Receding Hairline}{8} & \head{Wearing Necktie}{7} & \head{Rosy Cheeks}{6} & \head{Eyeglasses}{6}\\
  \icu{5_o_Clock_Shadow} & \icu{Receding_Hairline} & \icu{Wearing_Necktie} & \icu{Rosy_Cheeks} & \icu{Eyeglasses} \\
  \icb{5_o_Clock_Shadow} & \icb{Receding_Hairline} & \icb{Wearing_Necktie} & \icb{Rosy_Cheeks} & \icb{Eyeglasses} \\[.5ex]

  \head{Goatee}{6} & \head{Chubby}{6} & \head{Sideburns}{6} & \head{Blurry}{5} & \head{Wearing Hat}{5}\\
  \icu{Goatee} & \icu{Chubby} & \icu{Sideburns} & \icu{Blurry} & \icu{Wearing_Hat} \\
  \icb{Goatee} & \icb{Chubby} & \icb{Sideburns} & \icb{Blurry} & \icb{Wearing_Hat} \\[.5ex]

  \head{Double Chin}{5} & \head{Pale Skin}{4} & \head{Gray Hair}{4} & \head{Mustache}{4} & \head{Bald}{2}\\
  \icu{Double_Chin} & \icu{Pale_Skin} & \icu{Gray_Hair} & \icu{Mustache} & \icu{Bald} \\
  \icb{Double_Chin} & \icb{Pale_Skin} & \icb{Gray_Hair} & \icb{Mustache} & \icb{Bald}
  \end{tabular}}

  \Caption[fig:grad-cam]{Averaged Grad-CAM Activations}{This figure displays the average CAM activations for 40 different attributes including the probability of positive label $p_m$. Activations are averaged across all negative (left) and positive (right) predictions, extracted by AFFACT-u (top) and AFFACT-b (bottom).}
  \end{figure*}

\begin{figure*} [p]
  \centering
  \setlength{\tabcolsep}{1.5pt}
  \renewcommand{\arraystretch}{0.5}
  \scalebox{.85}{
  \begin{tabular}{*{4}{c@{\,}c@{\hspace*{1em}}}c@{\,}c}
  \headgcamplus{Attractive}{51} & \headgcamplus{Mouth Slgt. Open}{48} & \headgcamplus{Smiling}{48} & \headgcamplus{Wearing Lipstick}{47} & \headgcamplus{High Cheekbones}{45}\\
  \icugcamplus{Attractive} & \icugcamplus{Mouth_Slightly_Open} & \icugcamplus{Smiling} & \icugcamplus{Wearing_Lipstick} & \icugcamplus{High_Cheekbones} \\
  \icbgcamplus{Attractive} & \icbgcamplus{Mouth_Slightly_Open} & \icbgcamplus{Smiling} & \icbgcamplus{Wearing_Lipstick} & \icbgcamplus{High_Cheekbones} \\[.5ex]

  \headgcamplus{Male}{42} & \headgcamplus{Heavy Makeup}{38} & \headgcamplus{Wavy Hair}{32} & \headgcamplus{Oval Face}{28} & \headgcamplus{Pointy Nose}{28}\\
  \icugcamplus{Male} & \icugcamplus{Heavy_Makeup} & \icugcamplus{Wavy_Hair} & \icugcamplus{Oval_Face} & \icugcamplus{Pointy_Nose} \\
  \icbgcamplus{Male} & \icbgcamplus{Heavy_Makeup} & \icbgcamplus{Wavy_Hair} & \icbgcamplus{Oval_Face} & \icbgcamplus{Pointy_Nose} \\[.5ex]

  \headgcamplus{Arched Eyebrows}{27} & \headgcamplus{Big Lips}{24} & \headgcamplus{Black Hair}{24} & \headgcamplus{Big Nose}{24} & \headgcamplus{Young}{78}\\
  \icugcamplus{Arched_Eyebrows} & \icugcamplus{Big_Lips} & \icugcamplus{Black_Hair} & \icugcamplus{Big_Nose} & \icugcamplus{Young} \\
  \icbgcamplus{Arched_Eyebrows} & \icbgcamplus{Big_Lips} & \icbgcamplus{Black_Hair} & \icbgcamplus{Big_Nose} & \icbgcamplus{Young} \\[.5ex]

  \headgcamplus{Straight Hair}{21} & \headgcamplus{Brown Hair}{20} & \headgcamplus{Bags Under Eyes}{20} & \headgcamplus{Wearing Earrings}{19} & \headgcamplus{No Beard}{83}\\
  \icugcamplus{Straight_Hair} & \icugcamplus{Brown_Hair} & \icugcamplus{Bags_Under_Eyes} & \icugcamplus{Wearing_Earrings} & \icugcamplus{No_Beard} \\
  \icbgcamplus{Straight_Hair} & \icbgcamplus{Brown_Hair} & \icbgcamplus{Bags_Under_Eyes} & \icbgcamplus{Wearing_Earrings} & \icbgcamplus{No_Beard} \\[.5ex]

  \headgcamplus{Bangs}{15} & \headgcamplus{Blond Hair}{15} & \headgcamplus{Bushy Eyebrows}{14} & \headgcamplus{Wearing Necklace}{12} & \headgcamplus{Narrow Eyes}{12}\\
  \icugcamplus{Bangs} & \icugcamplus{Blond_Hair} & \icugcamplus{Bushy_Eyebrows} & \icugcamplus{Wearing_Necklace} & \icugcamplus{Narrow_Eyes} \\
  \icbgcamplus{Bangs} & \icbgcamplus{Blond_Hair} & \icbgcamplus{Bushy_Eyebrows} & \icbgcamplus{Wearing_Necklace} & \icbgcamplus{Narrow_Eyes} \\[.5ex]

  \headgcamplus{5 o'Clock Shadow}{11} & \headgcamplus{Receding Hairline}{8} & \headgcamplus{Wearing Necktie}{7} & \headgcamplus{Rosy Cheeks}{6} & \headgcamplus{Eyeglasses}{6}\\
  \icugcamplus{5_o_Clock_Shadow} & \icugcamplus{Receding_Hairline} & \icugcamplus{Wearing_Necktie} & \icugcamplus{Rosy_Cheeks} & \icugcamplus{Eyeglasses} \\
  \icbgcamplus{5_o_Clock_Shadow} & \icbgcamplus{Receding_Hairline} & \icbgcamplus{Wearing_Necktie} & \icbgcamplus{Rosy_Cheeks} & \icbgcamplus{Eyeglasses} \\[.5ex]

  \headgcamplus{Goatee}{6} & \headgcamplus{Chubby}{6} & \headgcamplus{Sideburns}{6} & \headgcamplus{Blurry}{5} & \headgcamplus{Wearing Hat}{5}\\
  \icugcamplus{Goatee} & \icugcamplus{Chubby} & \icugcamplus{Sideburns} & \icugcamplus{Blurry} & \icugcamplus{Wearing_Hat} \\
  \icbgcamplus{Goatee} & \icbgcamplus{Chubby} & \icbgcamplus{Sideburns} & \icbgcamplus{Blurry} & \icbgcamplus{Wearing_Hat} \\[.5ex]

  \headgcamplus{Double Chin}{5} & \headgcamplus{Pale Skin}{4} & \headgcamplus{Gray Hair}{4} & \headgcamplus{Mustache}{4} & \headgcamplus{Bald}{2}\\
  \icugcamplus{Double_Chin} & \icugcamplus{Pale_Skin} & \icugcamplus{Gray_Hair} & \icugcamplus{Mustache} & \icugcamplus{Bald} \\
  \icbgcamplus{Double_Chin} & \icbgcamplus{Pale_Skin} & \icbgcamplus{Gray_Hair} & \icbgcamplus{Mustache} & \icbgcamplus{Bald}
  \end{tabular}}

  \Caption[fig:gcampluscam-activation]{Averaged GradCAM++ Activations}{This figure displays the average CAM activations for 40 different attributes including the probability of positive label $p_m$. Activations are averaged across all negative (left) and positive (right) predictions, extracted by AFFACT-u (top) and AFFACT-b (bottom).}
\end{figure*}

\begin{figure*} [p]
  \centering
  \setlength{\tabcolsep}{1.5pt}
  \renewcommand{\arraystretch}{0.5}
  \scalebox{.85}{
  \begin{tabular}{*{4}{c@{\,}c@{\hspace*{1em}}}c@{\,}c}
  \headhires{Attractive}{51} & \headhires{Mouth Slgt. Open}{48} & \headhires{Smiling}{48} & \headhires{Wearing Lipstick}{47} & \headhires{High Cheekbones}{45}\\
  \icuhires{Attractive} & \icuhires{Mouth_Slightly_Open} & \icuhires{Smiling} & \icuhires{Wearing_Lipstick} & \icuhires{High_Cheekbones} \\
  \icbhires{Attractive} & \icbhires{Mouth_Slightly_Open} & \icbhires{Smiling} & \icbhires{Wearing_Lipstick} & \icbhires{High_Cheekbones} \\[.5ex]

  \headhires{Male}{42} & \headhires{Heavy Makeup}{38} & \headhires{Wavy Hair}{32} & \headhires{Oval Face}{28} & \headhires{Pointy Nose}{28}\\
  \icuhires{Male} & \icuhires{Heavy_Makeup} & \icuhires{Wavy_Hair} & \icuhires{Oval_Face} & \icuhires{Pointy_Nose} \\
  \icbhires{Male} & \icbhires{Heavy_Makeup} & \icbhires{Wavy_Hair} & \icbhires{Oval_Face} & \icbhires{Pointy_Nose} \\[.5ex]

  \headhires{Arched Eyebrows}{27} & \headhires{Big Lips}{24} & \headhires{Black Hair}{24} & \headhires{Big Nose}{24} & \headhires{Young}{78}\\
  \icuhires{Arched_Eyebrows} & \icuhires{Big_Lips} & \icuhires{Black_Hair} & \icuhires{Big_Nose} & \icuhires{Young} \\
  \icbhires{Arched_Eyebrows} & \icbhires{Big_Lips} & \icbhires{Black_Hair} & \icbhires{Big_Nose} & \icbhires{Young} \\[.5ex]

  \headhires{Straight Hair}{21} & \headhires{Brown Hair}{20} & \headhires{Bags Under Eyes}{20} & \headhires{Wearing Earrings}{19} & \headhires{No Beard}{83}\\
  \icuhires{Straight_Hair} & \icuhires{Brown_Hair} & \icuhires{Bags_Under_Eyes} & \icuhires{Wearing_Earrings} & \icuhires{No_Beard} \\
  \icbhires{Straight_Hair} & \icbhires{Brown_Hair} & \icbhires{Bags_Under_Eyes} & \icbhires{Wearing_Earrings} & \icbhires{No_Beard} \\[.5ex]

  \headhires{Bangs}{15} & \headhires{Blond Hair}{15} & \headhires{Bushy Eyebrows}{14} & \headhires{Wearing Necklace}{12} & \headhires{Narrow Eyes}{12}\\
  \icuhires{Bangs} & \icuhires{Blond_Hair} & \icuhires{Bushy_Eyebrows} & \icuhires{Wearing_Necklace} & \icuhires{Narrow_Eyes} \\
  \icbhires{Bangs} & \icbhires{Blond_Hair} & \icbhires{Bushy_Eyebrows} & \icbhires{Wearing_Necklace} & \icbhires{Narrow_Eyes} \\[.5ex]

  \headhires{5 o'Clock Shadow}{11} & \headhires{Receding Hairline}{8} & \headhires{Wearing Necktie}{7} & \headhires{Rosy Cheeks}{6} & \headhires{Eyeglasses}{6}\\
  \icuhires{5_o_Clock_Shadow} & \icuhires{Receding_Hairline} & \icuhires{Wearing_Necktie} & \icuhires{Rosy_Cheeks} & \icuhires{Eyeglasses} \\
  \icbhires{5_o_Clock_Shadow} & \icbhires{Receding_Hairline} & \icbhires{Wearing_Necktie} & \icbhires{Rosy_Cheeks} & \icbhires{Eyeglasses} \\[.5ex]

  \headhires{Goatee}{6} & \headhires{Chubby}{6} & \headhires{Sideburns}{6} & \headhires{Blurry}{5} & \headhires{Wearing Hat}{5}\\
  \icuhires{Goatee} & \icuhires{Chubby} & \icuhires{Sideburns} & \icuhires{Blurry} & \icuhires{Wearing_Hat} \\
  \icbhires{Goatee} & \icbhires{Chubby} & \icbhires{Sideburns} & \icbhires{Blurry} & \icbhires{Wearing_Hat} \\[.5ex]

  \headhires{Double Chin}{5} & \headhires{Pale Skin}{4} & \headhires{Gray Hair}{4} & \headhires{Mustache}{4} & \headhires{Bald}{2}\\
  \icuhires{Double_Chin} & \icuhires{Pale_Skin} & \icuhires{Gray_Hair} & \icuhires{Mustache} & \icuhires{Bald} \\
  \icbhires{Double_Chin} & \icbhires{Pale_Skin} & \icbhires{Gray_Hair} & \icbhires{Mustache} & \icbhires{Bald}
  \end{tabular}}

  \Caption[fig:hirescam-activation]{Averaged HiResCAM Activations}{This figure displays the average CAM activations for 40 different attributes including the probability of positive label $p_m$. Activations are averaged across all negative (left) and positive (right) predictions, extracted by AFFACT-u (top) and AFFACT-b (bottom).}
\end{figure*}

\begin{figure*} [p]
  \centering
  \setlength{\tabcolsep}{1.5pt}
  \renewcommand{\arraystretch}{0.5}
  \scalebox{.85}{
  \begin{tabular}{*{4}{c@{\,}c@{\hspace*{1em}}}c@{\,}c}
  \headelementwise{Attractive}{51} & \headelementwise{Mouth Slgt. Open}{48} & \headelementwise{Smiling}{48} & \headelementwise{Wearing Lipstick}{47} & \headelementwise{High Cheekbones}{45}\\
  \icuelementwise{Attractive} & \icuelementwise{Mouth_Slightly_Open} & \icuelementwise{Smiling} & \icuelementwise{Wearing_Lipstick} & \icuelementwise{High_Cheekbones} \\
  \icbelementwise{Attractive} & \icbelementwise{Mouth_Slightly_Open} & \icbelementwise{Smiling} & \icbelementwise{Wearing_Lipstick} & \icbelementwise{High_Cheekbones} \\[.5ex]

  \headelementwise{Male}{42} & \headelementwise{Heavy Makeup}{38} & \headelementwise{Wavy Hair}{32} & \headelementwise{Oval Face}{28} & \headelementwise{Pointy Nose}{28}\\
  \icuelementwise{Male} & \icuelementwise{Heavy_Makeup} & \icuelementwise{Wavy_Hair} & \icuelementwise{Oval_Face} & \icuelementwise{Pointy_Nose} \\
  \icbelementwise{Male} & \icbelementwise{Heavy_Makeup} & \icbelementwise{Wavy_Hair} & \icbelementwise{Oval_Face} & \icbelementwise{Pointy_Nose} \\[.5ex]

  \headelementwise{Arched Eyebrows}{27} & \headelementwise{Big Lips}{24} & \headelementwise{Black Hair}{24} & \headelementwise{Big Nose}{24} & \headelementwise{Young}{78}\\
  \icuelementwise{Arched_Eyebrows} & \icuelementwise{Big_Lips} & \icuelementwise{Black_Hair} & \icuelementwise{Big_Nose} & \icuelementwise{Young} \\
  \icbelementwise{Arched_Eyebrows} & \icbelementwise{Big_Lips} & \icbelementwise{Black_Hair} & \icbelementwise{Big_Nose} & \icbelementwise{Young} \\[.5ex]

  \headelementwise{Straight Hair}{21} & \headelementwise{Brown Hair}{20} & \headelementwise{Bags Under Eyes}{20} & \headelementwise{Wearing Earrings}{19} & \headelementwise{No Beard}{83}\\
  \icuelementwise{Straight_Hair} & \icuelementwise{Brown_Hair} & \icuelementwise{Bags_Under_Eyes} & \icuelementwise{Wearing_Earrings} & \icuelementwise{No_Beard} \\
  \icbelementwise{Straight_Hair} & \icbelementwise{Brown_Hair} & \icbelementwise{Bags_Under_Eyes} & \icbelementwise{Wearing_Earrings} & \icbelementwise{No_Beard} \\[.5ex]

  \headelementwise{Bangs}{15} & \headelementwise{Blond Hair}{15} & \headelementwise{Bushy Eyebrows}{14} & \headelementwise{Wearing Necklace}{12} & \headelementwise{Narrow Eyes}{12}\\
  \icuelementwise{Bangs} & \icuelementwise{Blond_Hair} & \icuelementwise{Bushy_Eyebrows} & \icuelementwise{Wearing_Necklace} & \icuelementwise{Narrow_Eyes} \\
  \icbelementwise{Bangs} & \icbelementwise{Blond_Hair} & \icbelementwise{Bushy_Eyebrows} & \icbelementwise{Wearing_Necklace} & \icbelementwise{Narrow_Eyes} \\[.5ex]

  \headelementwise{5 o'Clock Shadow}{11} & \headelementwise{Receding Hairline}{8} & \headelementwise{Wearing Necktie}{7} & \headelementwise{Rosy Cheeks}{6} & \headelementwise{Eyeglasses}{6}\\
  \icuelementwise{5_o_Clock_Shadow} & \icuelementwise{Receding_Hairline} & \icuelementwise{Wearing_Necktie} & \icuelementwise{Rosy_Cheeks} & \icuelementwise{Eyeglasses} \\
  \icbelementwise{5_o_Clock_Shadow} & \icbelementwise{Receding_Hairline} & \icbelementwise{Wearing_Necktie} & \icbelementwise{Rosy_Cheeks} & \icbelementwise{Eyeglasses} \\[.5ex]

  \headelementwise{Goatee}{6} & \headelementwise{Chubby}{6} & \headelementwise{Sideburns}{6} & \headelementwise{Blurry}{5} & \headelementwise{Wearing Hat}{5}\\
  \icuelementwise{Goatee} & \icuelementwise{Chubby} & \icuelementwise{Sideburns} & \icuelementwise{Blurry} & \icuelementwise{Wearing_Hat} \\
  \icbelementwise{Goatee} & \icbelementwise{Chubby} & \icbelementwise{Sideburns} & \icbelementwise{Blurry} & \icbelementwise{Wearing_Hat} \\[.5ex]

  \headelementwise{Double Chin}{5} & \headelementwise{Pale Skin}{4} & \headelementwise{Gray Hair}{4} & \headelementwise{Mustache}{4} & \headelementwise{Bald}{2}\\
  \icuelementwise{Double_Chin} & \icuelementwise{Pale_Skin} & \icuelementwise{Gray_Hair} & \icuelementwise{Mustache} & \icuelementwise{Bald} \\
  \icbelementwise{Double_Chin} & \icbelementwise{Pale_Skin} & \icbelementwise{Gray_Hair} & \icbelementwise{Mustache} & \icbelementwise{Bald}
  \end{tabular}}

  \Caption[fig:elementwisecam-activation]{Averaged Element-Wise CAM Activations}{This figure displays the average CAM activations for 40 different attributes including the probability of positive label $p_m$. Activations are averaged across all negative (left) and positive (right) predictions, extracted by AFFACT-u (top) and AFFACT-b (bottom).}
\end{figure*}

\end{onecolumn}
\fi

\end{document}